# Consequences of Slow Neural Dynamics for Incremental Learning

**Shima Rhaimi Moghaddam, Fanjun Bu†, Christopher J. Honey**

Johns Hopkins University, Baltimore, MD, USA.

Correspondence to: sh.rahimi.m@gmail.com

## Abstract

In the human brain, internal states are often correlated over time (due to local recurrence and other intrinsic circuit properties), punctuated by abrupt transitions. At first glance, temporal smoothness of internal states presents a problem for learning input-output mappings (e.g. category labels for images), because the internal representation of the input will contain a mixture of current input and prior inputs. However, when training with naturalistic data (e.g. movies) there is also temporal autocorrelation in the input. How does the temporal "smoothness" of internal states affect the efficiency of learning when the training data are also temporally smooth? How does it affect the kinds of representations that are learned? We found that, when trained with temporally smooth data, "slow" neural networks (equipped with linear recurrence and gating mechanisms) learned to categorize more efficiently than feedforward networks. Furthermore, networks with linear recurrence and multi-timescale gating could learn internal representations that "un-mixed" quickly-varying and slowly-varying data sources. Together, these findings demonstrate how a fundamental property of cortical dynamics (their temporal autocorrelation) can serve as an inductive bias, leading to more efficient category learning and to the representational separation of fast and slow sources in the environment.

## 1. Introduction

Organized artificial neural networks are thought to model how cortical hierarchies learn and represent information (Grill-Spector and Weiner 2013; Nayebi et al. 2018; Spoerer, McClure, and Kriegeskorte 2017; Yamins and DiCarlo 2016). At the level of representation, the neural networks that are trained to perform visual classification provide our best models of population-level and neuron-level responses to real-world stimuli (Kell et al. 2018; Nayebi et al. 2018; Yamins et al. 2014; Yamins and DiCarlo 2016). And at the level of learning, there are now proposals for how backpropagation-like learning can be implemented in cortical circuits (Bartunov et al. 2018; Bengio et al. 2016; Lillicrap et al. 2016). However, circuits in the biological brain face additional challenges and constraints, above and beyond those faced by artificial neural networks. Consider the setting of supervised category learning, in which the learner is presented with a sequence of training samples from different categories, and must learn to accurately predict the category label of each sample. For example, suppose a person stops to watch a basketball game at the park, and they want to learn to recognize each player and associate them with the names shown on their jerseys. In theory, one could learn the names of the players by associating the internal state associated with an input (e.g. the visual representation of a basketball player) with a categorical feature (e.g. the basketball player's name). However, there is a problem faced by cortical circuits performing this task: their dynamics exhibit *intrinsic autocorrelation* on the scale of milliseconds to seconds (Bright et al. 2020; Chaudhuri et al. 2015; Honey et al. 2012; Kiebel, Daunizeau, and Friston 2008; Kietzmann et al. 2019; Murray et al. 2014; Smith et al. 2013). The intrinsic temporal autocorrelation of cortical circuits is observed even in the absence of external stimuli and is not simply a result of variations in higher-order inter-regional information flow (Burt et al. 2019; Chaudhuri et al. 2015; Gao et al. 2020). Therefore, even if the sequence of training samples (e.g. images of each player) are uncorrelated in time, the internal brain state associated with one training sample will not be independent from the brain state associated with the next training sample. Loosely speaking, the representation of one basketball player is mixed with the representation of another player seen a second earlier. This is a problem because most learning algorithms, such as backpropagation, make use of these internal states to determine how to update the network. Moreover, algorithms which can handle high-dimensional temporal credit assignment, such as backpropagation through time, are generally not thought to be biologically plausible (Lillicrap et al. 2020; Lillicrap and Santoro 2019; Whittington and Bogacz 2019). Thus, it appears that the temporal autocorrelation intrinsic to neural dynamics will affect how biological brains can learn incrementally from sequences of examples.

† Work done while at Johns Hopkins University. Current address: Cornell Tech, New York City, NY, USA.

Our central research question is: *how are cortical circuits able to learn in the face of intrinsic temporal autocorrelation?* To address this question, we set out to investigate how conventional backpropagation learning is affected when there is autocorrelation in the internal states of the system, and the learning algorithm is unable to correct for this autocorrelation.

We hypothesized that there are two features of real-world learning which may reduce the costs (and possibly generate benefits) of the intrinsic autocorrelation in cortical circuits. First, although cortical circuits exhibit slow dynamics which manifest as temporal autocorrelation, they can also shift state suddenly at *event boundaries*, a process that may be associated with "resetting" of context representations (Baldassano et al. 2017; Chien and Honey 2020; Dominey 2021; DuBrow and Davachi 2016). Thus, it may be possible for cortical circuits to minimize the costs of unwanted temporal interference by occasionally "resetting" slow internal dynamics in the learning system. Second, real-world training data is itself temporally autocorrelated, as the samples we acquire from the world at one moment in time will tend to resemble those we obtain at the next moment (Hénaff, Goris, and Simoncelli 2019). If the training data are temporally autocorrelated, then perhaps the autocorrelation in internal dynamics can in fact serve as a useful inductive bias, consistent with properties of the environment. Moreover, data sampled from a slowly- changing environment may contain important features that are stable over time, and such stable features may be more readily extracted by a circuit that smoothly combines training data features over time.

To quantify the influences of intrinsic temporal autocorrelation of neural dynamics, and determine how it interacts with autocorrelation in the data, and with occasional resetting, we conducted a series of computational experiments that manipulated each of these factors. In particular, we examined how neural network learning was affected by introducing: (i) local linear recurrence in internal representations; (ii) a gating mechanism that resets internal representation at transitions between categories.

In particular, we asked the following questions:

i. How is supervised category learning affected when the network is equipped with (i) local linear recurrence in internal representations and (ii) a gating mechanism that resets internal representation at category transitions?

   *We hypothesized that when there is autocorrelation in data, networks with local linear recurrence and gating could exploit temporal autocorrelation in data for more efficient category learning by integrating and combining temporally correlated data while avoiding interference from uncorrelated data.*

ii. How do these local linear recurrence and gating mechanisms influence the internal representations that are learned from data with multi-timescale autocorrelation?

   *We hypothesized that networks with linear recurrence and gating could exploit multi-timescale temporal autocorrelation in data to generate separated internal representations of the quickly-changing and slowly-changing components of the training data.*

# 2. Results

## 2.1. Learning categories from temporally correlated data

We first explored whether brain-inspired recurrence and gating mechanisms can enable neural networks to exploit temporally correlated data for higher learning efficiency in classification tasks. In particular, we tested the ability of these models to perform supervised category learning in an incremental learning setting, in which the models were updated after each training sample (Fig 1 A). We tested learning on multiple datasets including MNIST, Fashion-MNIST, and synthetic datasets in which we could control the feature relationships between consecutive training items.

We manipulated the amount of temporal autocorrelation in the data by varying the number of consecutive samples drawn from the same category (Fig 1B). We then compared performance across 3 learning architectures: a feedforward neural network (Fig 1C); a neural network with local linear recurrence in its internal representations (Fig 1D); and a neural network with local linear recurrence and gating in its internal representations (Fig 1E). Importantly, in the networks with local linear recurrence, the recurrence is only within each unit, so that each unit linearly mixes its current input with its own past state (see schematic models in Fig 1D and E, and see Methods). Moreover, both feedforward and recurrent architectures were trained using conventional backpropagation: all models were trained as if they were feedforward networks, without propagating gradients backward in time.

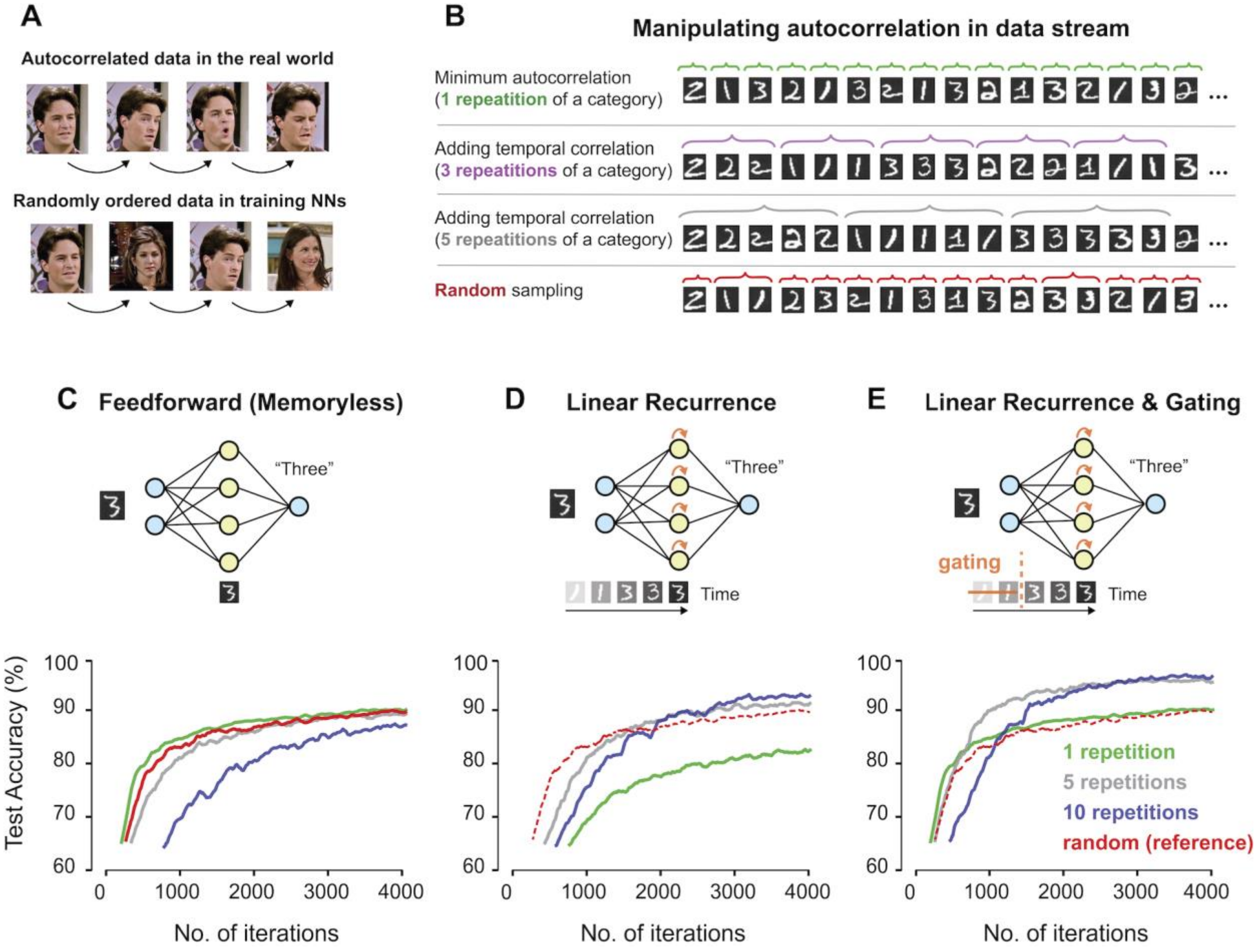

**Fig 1. Neural architectures for learning categories from temporally correlated data.** A) Top: temporally autocorrelated data in the real world. Bottom: randomly ordered data as the common practice in training neural networks. Photos in this section are taken from FRIENDS TV series, Warner Brothers (Kauffman et la, 1994). B) Manipulating autocorrelation levels in training data using the ordering of training samples. Colored rectangles indicate the amount of autocorrelation induced by repeating a category. C) Test accuracy in SGD training of a feedforward neural network (MNIST data) across different autocorrelation levels. D) Test accuracy for a neural network with linear recurrence in internal representations. E) Test accuracy for a neural network with linear recurrence and gating mechanisms. The performance of the feedforward neural network with random sampling (shown in red) is used as an identical reference in all 3 plots so it can be used for comparing performance across models. [Curves in this figure have been averaged over 5 runs with different initialization and were further smoothed using a 100-iteration moving average.]

In feedforward (memoryless) networks, learning was slower when training data were autocorrelated than when they were randomly and independently sampled (Fig 1C). Moreover, in these feedforward networks, minimum autocorrelation (i.e. avoiding sequential samples from the same category) yielded more efficient learning than random sampling (Fig 1C). These observations generalized across all tested datasets and across Mean Squared Error and Cross Entropy loss, with and without RMSprop optimization.

The superiority of minimum autocorrelation over other conditions in memoryless networks suggests that any level of correlation between consecutive training samples slows incremental learning, even the correlation that can occur by chance in random sampling (Fig 1C). Therefore, given a fixed time budget for training and a feedforward network, a sampling strategy that minimizes correlation can reach a higher performance than random sampling.

In contrast to the detrimental effects of autocorrelated data on feedforward (memoryless) learners, networks with simple linear recurrence learned more efficiently from autocorrelated training data (Fig 1D). Moreover, adding a gating mechanism to the recurrent units further accelerated and increased learning (Fig 1E). In learners with linear recurrence and gating, learning for all levels of autocorrelation in data was significantly faster than for random data and for data with minimum autocorrelation (avoiding sequential repetitions) (Fig 1E). These findings generalized across MNIST, Fashion-MNIST, and synthetic datasets.

### 2.1.1. Interim discussion

When data sampled at a given moment shares category-relevant features with recent samples, learners with simple linear recurrence, a form of leaky memory, were able to exploit this property for more efficient category learning (Fig 1D and E). Importantly, the resetting mechanism prevented the mixing of hidden representations from samples of different categories, allowing the system to benefit most from the correlation in data, while not suffering from between-category interference.

*Why does averaging of current and prior states produce more efficient learning from sequentially correlated data streams?*

The simple linear recurrence mechanism employed in our recurrent models acts like a leaky memory: for each unit in the network, the present input is linearly mixed with a representation of the unit's prior activation state. This linear mixing can be understood as a form of simple averaging or smoothing in time. Our working hypothesis is that averaging across consecutive members of the same category instantiates an assumption (an inductive bias) that consecutive samples from the data stream share task-relevant local features. When this assumption is satisfied, averaging (smoothing) across consecutive samples increases the proportion of variance in the hidden units that is associated with category-diagnostic features.

Our interpretation of the linear recurrence advantage (i.e. that it arises from successful averaging of related features) implies that if consecutive items in the data stream do not share any local features, then the benefits of recurrence will be eliminated. To test this, we trained our recurrent model on a synthetic data stream in which the consecutive items did not share local features (Fig 2). To eliminate shared local features between consecutive samples, we synthesized a data stream such that feature *i* from sample *(n)* was orthogonal (statistically independent and anticorrelated) to feature *i* from sample *(n+1)* (Fig 2). Consistent with our prediction, the recurrence advantage was eliminated in this setting, and linear recurrence, with or without gating, always learned more slowly than feedforward models (Fig 2).

Additionally, biological systems might employ this simple linear recurrence because, while it does not impair learning from random data streams, it can reduce the noise in the presence of temporal autocorrelation (S Fig. 1) and can enhance the accuracy of the categorization decision once multiple members of the same category have been seen sequentially (S Fig. 2).

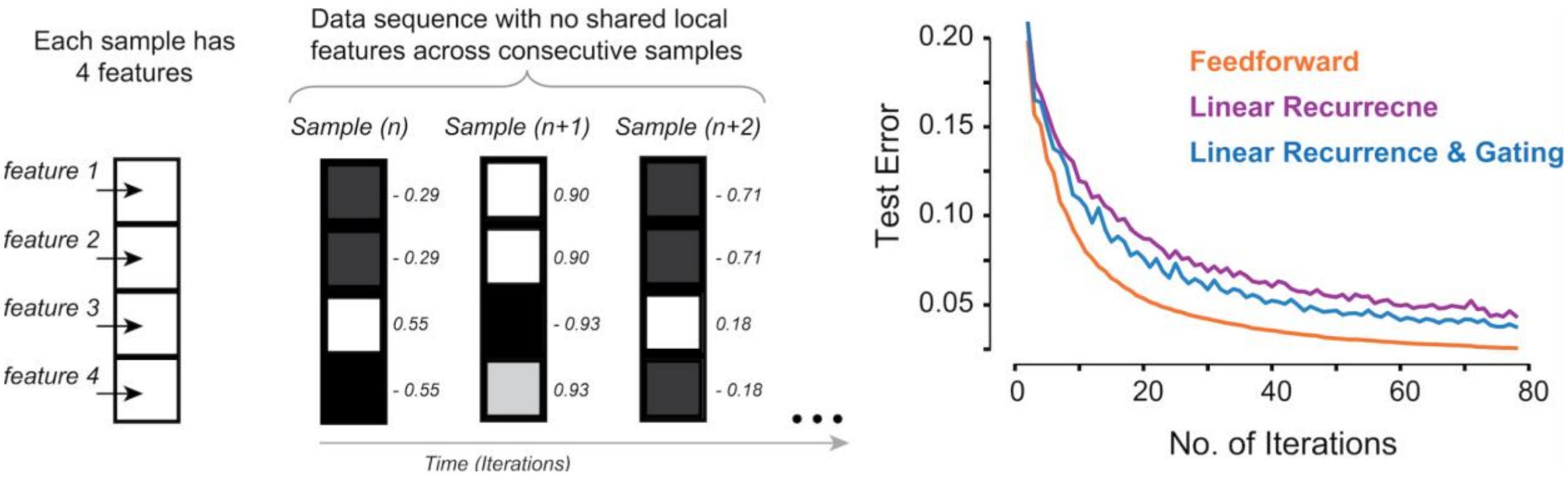

**Fig 2. Conditions under which recurrence and gating mechanisms impede category learning.** Left: Synthesized data stream with no shared local features across consecutive samples. Feature i in sample (n) is orthogonal to feature i in sample (n+1). The numbers on the right side of each sample show the feature values. Right: Test error for feedforward memoryless model, model with linear recurrence, and model with linear recurrence and gating.

*How does the model equipped with linear recurrence and gating compare with a more flexible recurrent model such as LSTM?*

The simple linear-recurrence-plus-gating model (Fig 1E) only allows for local averaging of past and present states, and is trained without backpropagating gradients through time (BPTT). It is important to compare the performance of this model to a more flexible recurrent architecture that can directly learn from task-relevant temporal structure using BPTT. In a preliminary analysis, we found that an LSTM trained with BPTT was able to benefit from training with autocorrelated data, learning more slowly at first, but ultimately achieving the lowest test error of all models (S Fig. 3 and S Fig. 4). However, the performance advantage of the LSTM trained with BPTT was not preserved when the models were tested out-of-domain. In particular, when models were trained on data that contained temporal autocorrelation, but tested on data with minimum autocorrelation (1 repetition per category), the linear-recurrence-plus-gating model showed better performance than LSTM models (Fig 3). We interpret these results as evidence that the LSTM has a much more flexible architecture, and via BPTT it can be calibrated to the structure of the training data stream (e.g. the LSTM can learn that there are precisely 5 repetitions in a block). Conversely, linear-recurrence-plus-gating model is more biologically plausible, it is trained without BPTT, it showed categorization performance competitive with the LSTM, and it generalized better across different levels of temporal autocorrelation. Note that we obtained competitive performance between LSTM and linear-recurrence-plus-gating model even though the LSTM's gradient updates are mathematically optimized for the task (via BPTT), whereas the gradient computations in the linear-recurrence-plus-gating model do not account for the network recurrence at all.

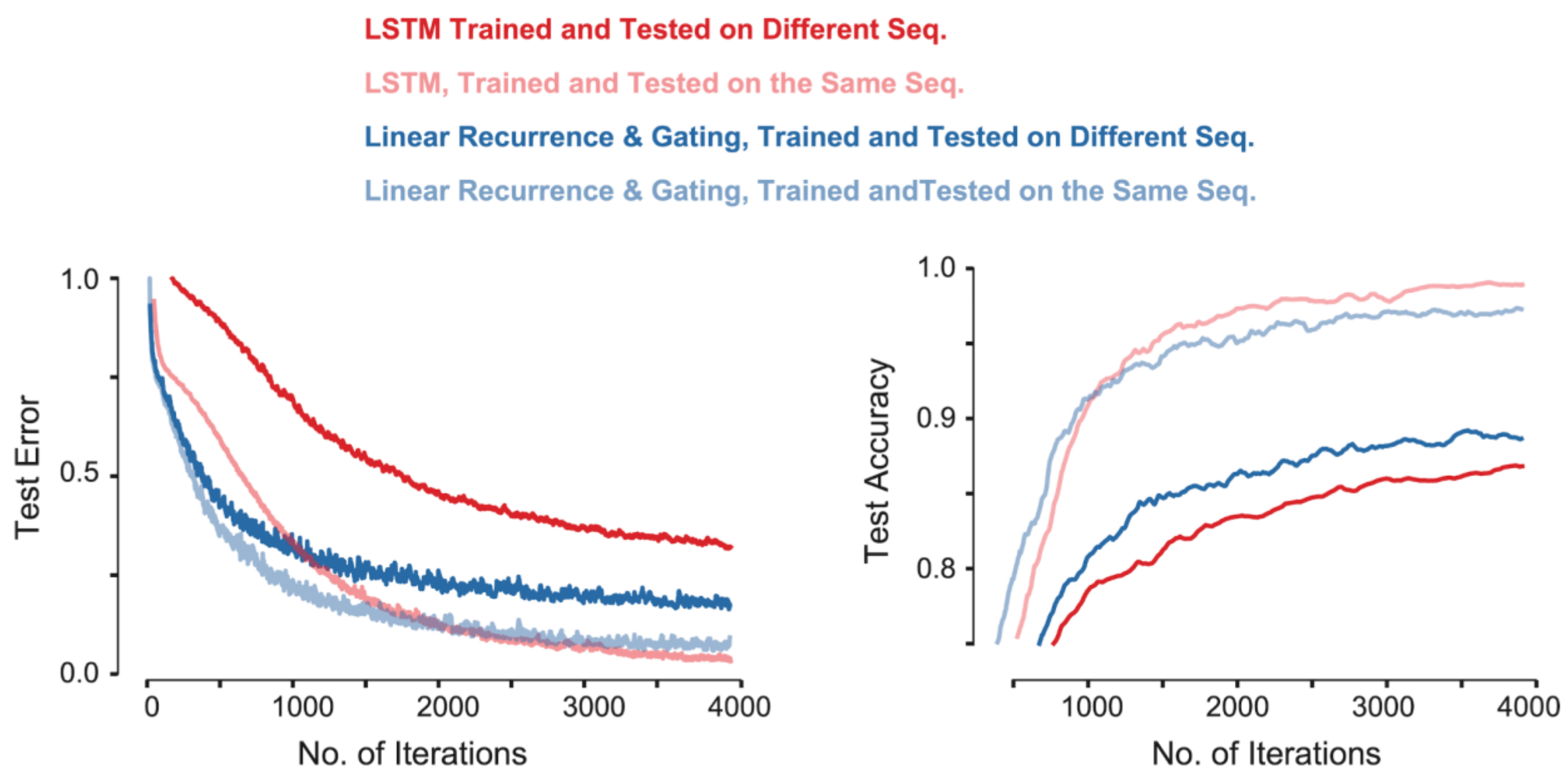

**Fig 3. Generalization of LSTM and linear-recurrence-plus-gating model to data streams with different temporal structure.** Left) Test error for LSTM and linear-recurrence-plus-gating model, trained and tested on the same sequence of samples or on a different sequence of samples. Right) Test accuracy for LSTM and linear-recurrence-plus-gating model, trained and tested on the same sequence of samples or on a different sequence of samples.

*Are the benefits of linear recurrence due to a form of gradient averaging, analogous to mini-batching?*

Networks with linear recurrence average *activations* over time, while batching averages *gradients* over time (over multiple samples). It appears that the two mechanisms are different, because the linear recurrence effects were reversed when the training categories contained non-overlapping features (Fig 2) and because recurrence and mini-batching affect performance in different ways as a function of the amount of category repetition (see S Fig. 5).

## 2.2. Learning temporally un-mixed representations from multi-scale autocorrelated data

In the real world, we may need to learn from data with multiple levels of autocorrelation. For instance, when having a face-to-face conversation: the features around a person's mouth change quickly, while their face's outline changes more slowly (Fig 4A). Moreover, there are no predefined labels to support the learning of representations in this setting. How does intrinsic temporal autocorrelation affect representation learning in this multi-scale setting? To test this, we characterized the internal representations learned by networks with linear recurrence mechanisms and gating mechanisms.

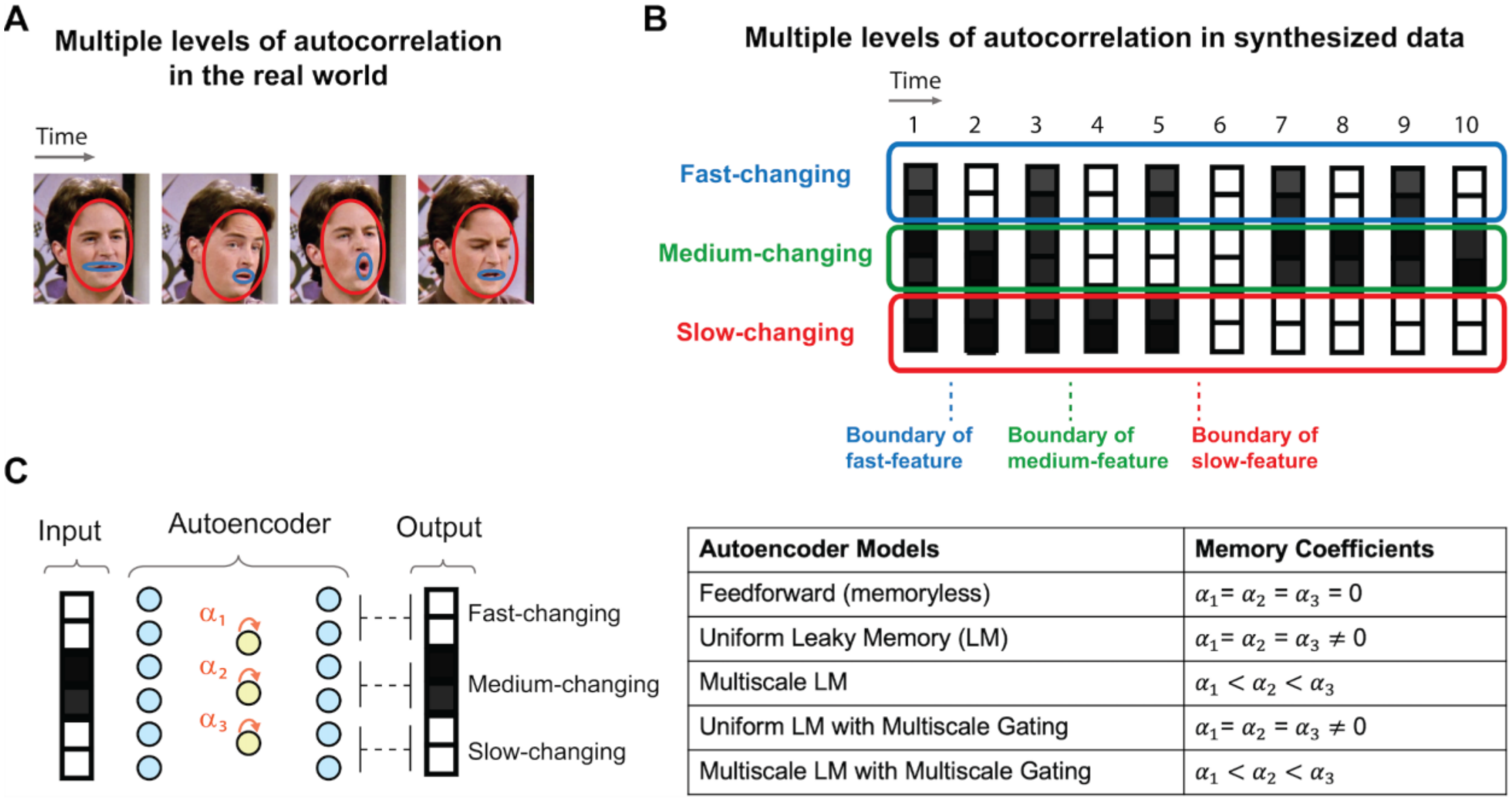

| Autoencoder Models | Memory Coefficients |
|---|---|
| Feedforward (memoryless) | $\alpha_1 = \alpha_2 = \alpha_3 = 0$ |
| Uniform Leaky Memory (LM) | $\alpha_1 = \alpha_2 = \alpha_3 \neq 0$ |
| Multiscale LM | $\alpha_1 < \alpha_2 < \alpha_3$ |
| Uniform LM with Multiscale Gating | $\alpha_1 = \alpha_2 = \alpha_3 \neq 0$ |
| Multiscale LM with Multiscale Gating | $\alpha_1 < \alpha_2 < \alpha_3$ |

**Fig 4. Unsupervised learning from data with autocorrelation on multiple timescales.** A) Example of multiple levels of autocorrelation in samples from the real world: the mouth shape changes more quickly than face-shape, and so has shorter autocorrelation time. B) Multiple levels of autocorrelation in synthesized data: first 2 elements change every item, representing the fast-changing feature; second two elements change every 3 items, representing a medium-changing feature; and third two elements change every 5 items, representing a slow-changing feature. X-axis shows time, each 6-by-1 item is one sample. C) 5 Different AE models: feedforward, uniform leaky memory (LM), multiscale LM, uniform LM with multiscale gating, and multiscale LM with multiscale gating. $\alpha_1$, $\alpha_2$, and $\alpha_3$ indicate the memory coefficient in the hidden representations (see Eq. 1).

We hypothesized that the timescales of temporal integration inside the learning network could interact with the timescales of autocorrelation in the data stream. Consistent with our broader theory that intrinsic temporal correlation serves as an inductive bias, we hypothesized that slowly (or infrequently) changing internal states in a simple recurrent architecture would come to represent more slowly varying features of the training data.

To test this, we examined whether neural networks equipped with multi-scale (i.e. fast and slow) linear recurrence and multiscale gating could learn separated representations of data structures that vary on fast and slow timescales. To evaluate such a timescale un-mixing ability across different architectures, we used a synthetic dataset whose temporal structure could be precisely controlled. In the training data, each sample contained 3 subcomponents, and these subcomponents changed at 3 different timescales (Fig 4B). We tested 5 different autoencoder (AE) architectures in their ability to reconstruct samples from the training data stream (Fig 4C). We varied which levels of recurrence and gating were available in each autoencoder model (see Methods for details). First, we test a feedforward AE. Second, we tested an AE with uniform leaky memory (LM). Third, we tested an AE with multiscale LM, in which some units in the model exhibit longer time constants in their leak memories. Fourth, we tested an AE with uniform LM and multiscale gating. In this model, all time constants were the same, but the memory resetting happens at different rates across different units. Finally, we tested an AE with both multiscale LM and multiscale gating.

We first confirmed that all the AE models could learn to reconstruct the input. The most efficient architectures (architectures with lowest test error) were the networks with LM and multiscale gating AE, and the feedforward memoryless AE (See test error in Fig 5D).

Networks equipped with both recurrence and multiscale gating could successfully un-mix the data, learning separated representations for the quickly and slowly changing features within the data stream (See "AE with Memory and Multiscale Gating" in Fig 5C). The individual hidden state units in these AE models were selectively more correlated with their corresponding data features (e.g. the slow-changing feature was more correlated with the long-memory node than with the other nodes; Fig 5C). These findings generalized across different synthesized datasets, different learning rates, and different recurrence (memory) coefficients (S Fig. 7).

We found that models with recurrence and multiscale gating showed significantly higher timescale selectivity (Fig 5D). Overall, the AE models with recurrence showed higher loss compared to feedforward AE (Fig 5D). However, in the models that had both recurrence and multiscale gating, the small increase in the loss was accompanied by a significant increase in timescale-selectivity scores (Fig 5D).

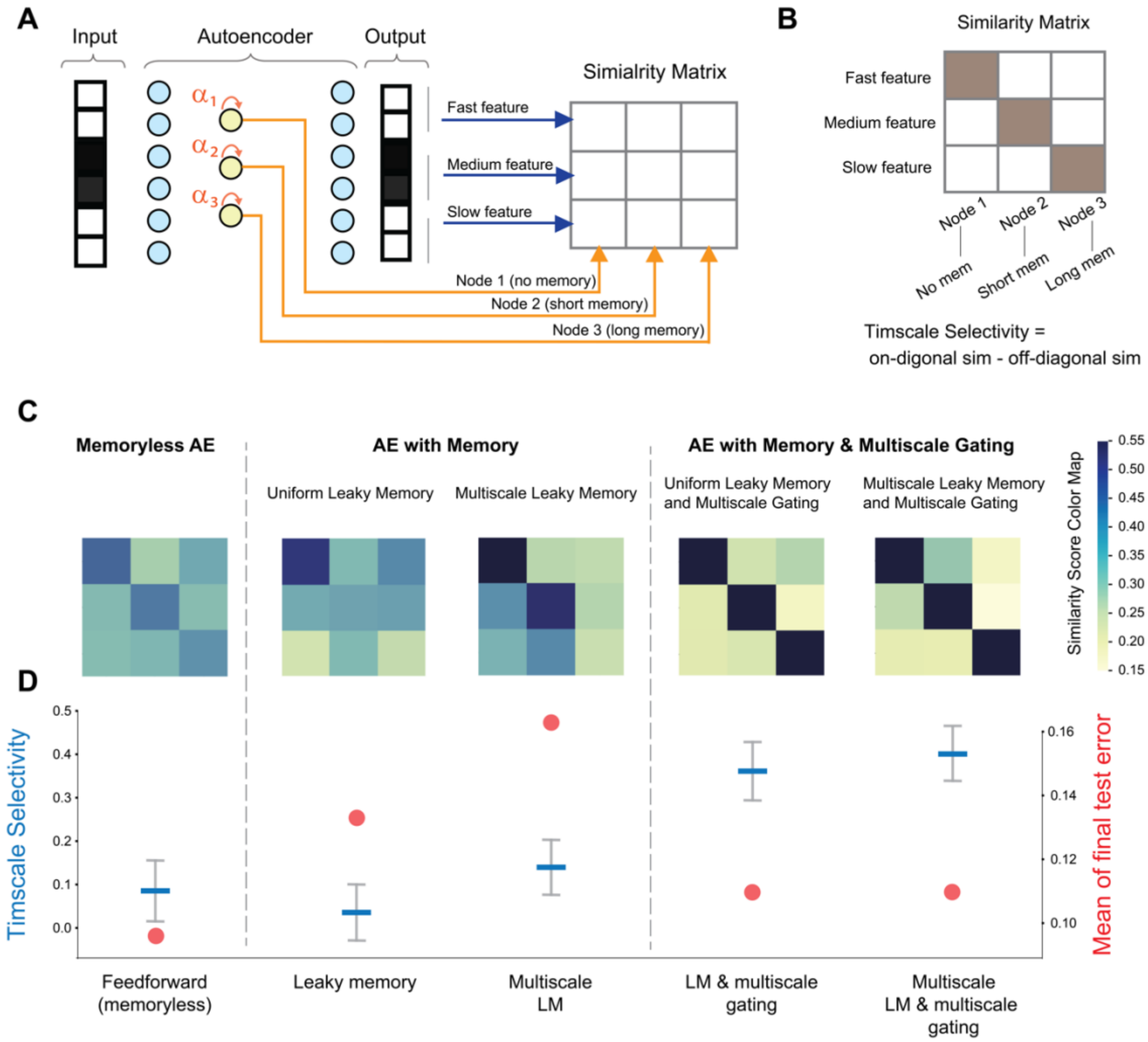

**Fig 5. Quantifying the similarity between internal representations and data features that vary on fast, medium, and slow timescales.** A) Illustration of which network-units and which training-data elements contribute to each element of the similarity matrix. For instance, the top-left element of the matrix shows the similarity between the fast feature of the data and the no-memory internal representation (node 1). B) Illustration of how timescale selectivity is computed. C) Similarity matrices for 3 groups of models: feedforward AE, AE with memory, and AE with memory and multiscale gating. D) Timescale selectivity and final test error for different AE models. Blue bars show the timescale selectivity. Each bar shows the 95% confidence interval for results from 40 runs with different random initializations. Red dots show the mean final test error across 40 runs. The memory coefficients of the models shown in (C) and (D) are as follows: leaky memory: 0.6; Multiscale LM: 0, 0.3, 0.9; LM & multiscale gating: 0.6; Multiscale LM and multiscale gating: 0, 0.3, 0.9.

### 2.2.1. Interim discussion

*Why might recurrence and gating mechanisms benefit the learning of timescale-separated representations?* The two autoencoder models that had both recurrence and multiscale gating mechanisms were most successful in learning internal representations that tracked distinct timescales of the input. Slowly (or quickly) varying features were extracted by slowly (or quickly) varying subsets of the network, analogous to a matched filter (see also (Mozer et al. 1992)). Features that change on different timescales may correspond to different levels of structure in the world (Wiskott and Sejnowski 2002). Thus, by adding recurrence and multiscale gating to a simple feedforward AE model, we equipped it with an ability to separate different levels of structure in the environment. Moreover, because intrinsic dynamics vary on multiple scales in the human brain (Honey et al. 2012; Murray et al. 2014; Raut, Snyder, and Raichle 2020; Smith et al. 2013) this implies that slowly-varying brain circuits may be biased to extract slowly-varying structure from the world (Honey, Newman, and Schapiro 2017).

*Why did the feedforward (memoryless) model produce slightly lower reconstruction error than models with recurrence?*

There is a cost to using AE models with recurrence (rather than feedforward AE models) to reconstruct data that varies on multiple timescales: overall test error increases slightly, most likely because slowly-changing internal states are ineffective for reconstructing quickly-changing features. In other words, we attribute this small additional error to the challenge of reconstructing input from an internal representation that is mismatched to the data. Consistent with this interpretation, we found that if a model's "slow" hidden units (i.e. medium and long memory units) were correlated with the fast-changing features in the training data, the model's per-feature error was worse (see S Fig. 6). This slight increase in reconstruction error is, however, accompanied by a significant benefit: learning more temporally interpretable, un-mixed representations of a multi-scale data stream.

# 3. Conclusions and general discussions

Primate brains' dynamics exhibit short- and long-range autocorrelation in their cortical dynamics (Bright et al. 2020; Chaudhuri et al. 2015; Honey et al. 2012; Murray et al. 2014; Raut, Snyder, and Raichle 2020; Smith et al. 2013). Initially, the temporal smoothness of internal states may seem problematic for learning input-output mappings such as image categorization, because a learning system must form associations between an output at one moment (e.g. a category label) and a combination of current and previous inputs (e.g. temporally smoothed internal representations). However, we noticed that two factors in real-world learning may reduce the costs and generate benefits of intrinsic autocorrelation in cortical circuits. First, although cortical circuits exhibit slow dynamics, they can shift state suddenly at event boundaries, minimizing the costs of unwanted temporal interference (Baldassano et al. 2017; Chien and Honey 2020; DuBrow and Davachi 2016). Second, real-world training data is itself temporally autocorrelated, which means the autocorrelation in internal dynamics might serve as an inductive bias that is consistent with properties of the environment. Therefore, to understand the consequences of intrinsic temporal autocorrelation, we conducted a series of computational experiments with networks forced to generate temporally smooth internal states punctuated by rapid transitions.

First, we examined the speed of learning in the setting of category learning. We found that temporally correlated training data slowed learning in feedforward neural networks, but sped learning in systems with linear recurrence (Fig 1C, D, E). Moreover, adding a simple gating mechanism to linearly recurrent networks enabled them to flexibly adapt to the autocorrelation level, so that they could benefit from repeating structure while not suffering from the interference of unrelated prior information.
Second, we examined the internal representations learned by neural networks performing unsupervised (reconstruction-based) learning on data streams containing both fast and slow temporal structure (Fig 4). Neural networks with recurrence and feature-sensitive gating learned representations that un-mixed features varying on different timescales (Fig 5). If distinct timescales in data reflect distinct data generators, these "un-mixed" representations may provide a more "meaningful" description of the input data (Jain et al. 2020; Mahto et al. 2020; Mitchell 2020).

For feedforward networks, it may be advantageous to train with minimum autocorrelation (i.e. avoiding consecutive samples from the same category) because minimal autocorrelation reduces the representational overlap across consecutive training items. Catastrophic forgetting can be reduced by decreasing the overlap between learned representations, for example, via orthogonalization (Robert M. French 1999). Though we did not explicitly seek to reduce interference by sampling with minimum autocorrelation, this method does likely reduce the representational overlap of nearby items. In addition, training with minimum autocorrelation may improve learning by maintaining a near-uniform distribution of sampled categories. Training with "low-discrepancy" sequences, such as those with uniformly distributed data, avoids classification bias and enhances learning (Iwata and Ishii 2002; Mishra and Rusch 2021).

The local linear recurrent networks we tested exhibited more efficient learning and more interpretable internal representations, even though they were trained with a learning rule that did not employ any temporal information. Specifically, all networks were trained incrementally using backpropagation with a loss function that only depended on the immediate state of the network. Architectures with recurrence and gating can thus exploit temporal structure in a way that is computationally simpler and more biologically plausible than backpropagation through time (Ilya Sutskever 2013; Lillicrap and Santoro 2019). With respect to biological plausibility, we note that the linear-recurrence-plus-gating system works well even for autoencoders (Fig 5), for which there are simple activation-based learning rules that do not require the propagation of partial derivatives (Lee et al. 2015). On the computational side, we highlight that the gradients computed for the recurrent networks were, in a sense "inaccurate", because the update rule was unaware of the recurrent leak connections, and yet the overall classification accuracy in the recurrent nets was higher than in feedforward nets, for which the gradients should be more accurate.

While prior work has focused on the implications of slow neural dynamics for online comprehension (Hasson, Chen, and Honey 2015) as well as working memory and decision making (Soltani et al. 2021), here we emphasize the implications of slow intrinsic dynamics for representation learning. At first glance, the presence of ubiquitous autocorrelation presents a problem for learning representations of individual items in the world: the information about one stimulus is mixed with information about stimuli that came before, and it is unclear how to resolve this interference without propagating gradients through time. However, here we demonstrate a beneficial effect: when the properties of the training data themselves are autocorrelated in time, then an intrinsic autocorrelation in neural dynamics can provide an inductive bias which accelerates learning (Fig 1D). Moreover, some of the unwanted interference between unrelated items can be reduced if neural circuits are additionally equipped with a resetting mechanism (Fig 1E).

We highlight four directions for future extension of this work. First, it would be interesting to examine the efficacy of linear recurrence and gating mechanisms in CNN architectures trained on video segmentation sequences (Wang and Hu 2021). Second, future work should test whether these results generalize to larger architectures and more realistic datasets, and should include a broader search of the hyperparameter space. We expect that the present results do have some generality, because we used simple architectures, made few domain-specific assumptions, and demonstrated the phenomena across multiple datasets. We expect the method to work best for datasets in which important or diagnostic data features persist over consecutive samples. Third, our autoencoder models were trained to reconstruct the present stimuli. Future work could extend our autoencoder architecture to a predictive architecture so that the model would predict the next stimulus rather than the present one (Arora and Cai 2021). Fourth, it would also be interesting to investigate the broader consequences of learning with recurrence: for example, human internal representations of natural sensory input sequences appear to be smooth in time, in contrast to the representations of most feedforward nets (Hénaff, Goris, and Simoncelli 2019), and training with smooth data and recurrence could potentially reduce this difference.

In sum, we explored the computational consequences of the temporal autocorrelation that is intrinsic to biological brains. Focusing on the setting of learning, we showed that neural networks with linear recurrence and gating can learn more quickly from temporally correlated data. Moreover, such networks also generate internal representations that separate distinct timescales of the data. Both of these results are observed in networks that learn incrementally and without needing to propagate gradients in time.

All in all, we conclude that intrinsic temporal autocorrelation is not an insurmountable obstacle for learning in real world settings; instead, temporal autocorrelation can be understood as an inductive bias that is useful in a realistic setting of incremental learning from a sequence of correlated exemplars.

# 4. Methods

## 4.1. Learning categories from temporally correlated data

### Classifier networks with brain-inspired constraints

**Linear recurrence:** We added local linear recurrence, a form of leaky memory, to the internal representations (hidden units) by linearly mixing them across consecutive time points. Hidden unit activations were updated according to following function:

$$H(n) = \alpha H(n-1) + (1 - \alpha)\, ReLU(W_{IH}\, I(n)) \qquad \textbf{Eq. 1}$$

where H(n) is the state of the hidden units for trial n, I(n) is the state of the input units for trial n, $\alpha$ is the recurrence (memory) coefficient, $W_{IH}$ are the connections from the input layer to the hidden layer, and ReLU is a rectified linear activation. We set $\alpha$ = 0.5 in these experiments.

**Gating:** In order to reduce the interference between items from different categories in the linear recurrence, we employed a gating mechanism to reset memory at the transitions between categories. Therefore, if sample n was drawn from a category other than the category of sample n-1, then we set $\alpha$ = 0 in *Eq. 1* on that trial n (Fig 1E).

### Manipulating autocorrelation in training data

We manipulated the amount of autocorrelation in data by varying the number of consecutive samples drawn from the same category. We began each training session by generating a random "category order", which was a permutation of the numbers from 1 to N (e.g. the ordering in Fig 1B is 2-1-3). The same category order was used for all conditions in that training session.

To sample with minimum autocorrelation, we sampled exactly one exemplar from each category, before sampling from the next category in the category order (1 repeat) (Fig 1B). This condition is called "minimum autocorrelation" because all consecutive items were from different categories, and there were not more examples from a category until all other categories were sampled. We increased autocorrelation by increasing the number of consecutive samples drawn from each category (3 repeats and 5 repeats in Fig 1B). Finally, we also used the standard random sampling method, in which items were sampled at random, without replacement, from the training set (Fig 1B). The training set was identical across all conditions, as was the order in which samples were drawn from within a category (Fig 1B).

### Training neural networks for classification

**Dataset.** We tested MNIST, Fashion-MNIST, and synthetic datasets containing low category overlap (LeCun, Cortes, and Burges 2010; Xiao, Rasul, and Vollgraf 2017). For creating and testing the models, we used PyTorch (Paszke et al. 2019).

**Learning Rule.** For the learning rule, we used backpropagation, however the gradient computation did not account for the fact that the neurons were leaky. Therefore, the update rule in linear-recurrence-plus-gating model is different from the common update rule in recurrent models (e.g. LSTM). LSTM uses backpropagation through time (BPTT), which is implausible for biological settings. In learning with BPTT, the same neurons must store and retrieve their entire activation history (Whittington and Bogacz 2019). In contrast, in the linear-recurrence-plus-gating model, neurons only use local information from their most recent history. Therefore, it is computationally much simpler because it does not require to maintain the whole history and to compute the gradient relative to all that history.

Objective function. We used backpropagation with both mean squared error (MSE) and cross-entropy (CE) loss functions. The results reported here are using MSE, primarily for the ease of comparison with later reconstruction error measures in this manuscript. However, the same pattern was observed using CE loss. Also, it has been shown MSE loss provides comparable performance to commonly utilized classification models with CE loss function. To test incremental learning, we employed stochastic gradient descent (SGD), updating weights for each training sample.

**Optimization, Initialization, and Activation Function.** We tested the model both with and without RMSprop optimization, along with Xavier initialization (Glorot and Bengio 2010; Tieleman and Hinton 2012). We applied ReLU to hidden units and Softmax or Sigmoid to the output units.

**Hyperparameters.** For MNIST and Fashion-MNIST, we used a 3-layer fully connected network with (784, 392, 10) dimensions and a learning rate of 0.01. The learning rate was not tuned for a specific condition. We used the same learning rate across all conditions; only autocorrelation varied across conditions. To compensate for the potential advantage of a specific set of hyperparameters for a specific condition, we ran 5 runs, each with a different random weight initialization, and reported the averaged results. When RMSprop was implemented,  and  were set to 0.9 and 0.99, respectively.

## 4.2. Learning interpretable representations from multi-scale temporally autocorrelated data

### Multi-timescale autocorrelation in training data

**Dataset.** To test the un-mixing abilities of our networks, we synthesized simplified training datasets which contained three levels of temporal structure. The input to the model at each time point consisted of 3 subcomponents (top, middle, bottom), and each subcomponent had two elements. Each subcomponent was generated to express a different level of autocorrelation over time: for example, the top, middle and bottom rows changed feature-category every 1, 3 or 5, iterations, respectively (Fig 4B). The individual features sampled at each time were generated as the sum of (i) an underlying binary state variable (which would switch every 1, 3 or 5 iterations) and (ii) uniformly distributed noise. As a result, the model was provided with features that varied at 3 timescales: fast (top row), medium (middle row), and slow (bottom row).

## Autoencoder architectures with brain-inspired constraints

**Brain-inspired constraints.** We used the same brain-inspired mechanisms for unsupervised learning models: local linear recurrence and gating mechanisms.

To implement linear recurrence at multiple scales, we varied the time constants across the nodes in the hidden layer. Thus, the variable a in *Eq. 1* was set to 0 for "no-memory", 0.3 for "short-memory", and 0.6 for "long-memory" nodes.

We used a multi-timescale gating mechanism that was sensitive to multi-timescale changes in the input stream. For each timescale in the input, the gating mechanism would use information from current and previous input to decide to reset memory when the change passed a threshold (see S Eq. 1) (Chien and Honey 2020): the internal representation in short-memory unit reset when there was a shift in the medium-changing property of the input stream; and, the internal representation in long-memory unit reset when there was a shift in the slow-changing property of the input stream.

**Architecture.** To evaluate the effectiveness of the added mechanisms, we compared 5 types of autoencoder (AE) models (Fig 4C): i) Feedforward AE; ii) AE with uniform linear recurrence in internal representations; iii) AE with multi-scale linear recurrence in internal representations, inspired by evidence showing that levels of processing in the brain can integrate information at different timescales (Bright et al. 2020; Chaudhuri et al. 2015; Honey et al. 2012; Kiebel, Daunizeau, and Friston 2008), and that multiple timescales are present even within a single circuit (Bernacchia et al. 2011; Ulanovsky et al. 2004); iv) AE with uniform linear recurrence in internal representations and multiscale boundary-sensitive gating, motivated by the evidence showing that processing in cortical circuits are sensitive to event-boundaries and these boundaries can shift learned representations (Chien and Honey 2020; Dominey 2021; DuBrow and Davachi 2016); and (v) AE with multiscale linear recurrence in internal representations and multiscale boundary-sensitive gating.

## Training autoencoder networks

**Learning algorithm, optimization, and initialization.** Similar to the classification task, we used backpropagation, while the gradient computation did not account for the fact that the neurons were leaky. We used MSE loss, both with and without RMSprop optimization method, and Xavier initialization (Glorot and Bengio 2010; Tieleman and Hinton 2012). We applied ReLU and Sigmoid as activation functions for hidden and output units, respectively.

**Hyperparameters.** The networks were 3-layer, fully connected autoencoders with (6, 3, 6) dimension. Learning rate was 0.005. In cases where RMSprop was implemented, the beta-1 and beta-2 were set to 0.9 and 0.99. For linear recurrence in internal representations, we tested a range of memory coefficients (a in Eq. 1). For uniform linear recurrence, we tested 3 sets of memory coefficients: model1: $\alpha1=\alpha2=\alpha3=0.3$; model2: $\alpha1=\alpha2=\alpha3=0.6$; model3: $\alpha1=\alpha2=\alpha3=0.9$. For multiscale linear recurrence, we tested the following combination of memory coefficients: model1: $\alpha1=0$; $\alpha2=0.3$; $\alpha3=0.6$; model2: $\alpha1=0$; $\alpha2=0.6$; $\alpha3=0.9$; model3: $\alpha1=0$; $\alpha2=0.3$; $\alpha3=0.9$.

## Evaluation methods for autoencoder networks

**Measure of multiscale representation-learning.** We measured the network's ability to "un-mix" the timescales of its input. By un-mixing, we mean learning representations that selectively track distinct latent sources that generated features within each training sample. In particular, we tested whether different nodes in the internal representations of the network would track different temporal features of the data stream: one node would track the fast-changing features of the data, a second node would track the medium-changing features of the data, and the third node would track the slow-changing features of the data.

To this end, using squared of Pearson correlation, we measured the similarity between the internal representations (node 1, node 2, node 3) and the data features (fast, medium, and slowly changing) (Fig 5A). In the models with multi-timescale linear recurrence, node 1 has no memory, node 2 has short memory, and node 3 has long memory. For AE models that do not have a multiscale memory, when referring to no-memory, short-memory, and long-memory nodes, we are referring to the hidden nodes in those positions (Fig 5A).

In the similarity matrix, the diagonal elements show the similarity of hidden representation to their corresponding timescale of data feature (similarity of fast-feature and no-memory hidden node, similarity of medium-feature and shot-memory hidden node, and similarity of slow-feature and long-memory node). Off-diagonal elements of similarity matrix show the similarity of hidden representation to their non-matching timescale of data feature (e.g. similarity of fast-feature and long-memory hidden node) (Fig 5A and B).

We then quantified the "timescale-selectivity" — e.g. whether the slow-changing feature was more correlated with the long-memory node than other nodes (no-memory and short-memory nodes) (Fig 5B). We therefore calculated timescale-selectivity as the average of the difference between the on-diagonal and the off-diagonal elements of the similarity matrix (Fig 5B).

**Measure of learning efficiency.** Learning efficiency was measured using the reconstruction error of the test data, computed as the MSE across all 3 subcomponents of each data sample.

## References

Arora, Tushar, and Ming Bo Cai. 2021. "Learning To Perceive Objects By Prediction." (Svrhm).

Baldassano, Christopher et al. 2017. "Discovering Event Structure in Continuous Narrative Perception and Memory." *Neuron* 95(3): 709-721.e5. http://dx.doi.org/10.1016/j.neuron.2017.06.041.

Bartunov, Sergey et al. 2018. "Assessing the Scalability of Biologically-Motivated Deep Learning Algorithms and Architectures." *arXiv* (Nips). http://arxiv.org/abs/1807.04587.

Bengio, Yoshua et al. 2016. "Towards Biologically Plausible Deep Learning." *arXiv* 7(5): 803–9. http://arxiv.org/abs/1502.04156.

Bernacchia, Alberto, Hyojung Seo, Daeyeol Lee, and Xiao Jing Wang. 2011. "A Reservoir of Time Constants for Memory Traces in Cortical Neurons." *Nature Neuroscience* 14(3): 366–72.

Bright, Ian M. et al. 2020. "A Temporal Record of the Past with a Spectrum of Time Constants in the Monkey Entorhinal Cortex." *Proceedings of the National Academy of Sciences of the United States of America* 117(33): 20274–83.

Burt, Joshua B et al. 2019. "Hierarchy of Transcriptomic Specialization across Human Cortex Captured by Structural Neuroimaging Tpography." *Nature Neuroscience* 21(9): 1251–59.

Chaudhuri, Rishidev et al. 2015. "A Large-Scale Circuit Mechanism for Hierarchical Dynamical Processing in the Primate Cortex." *Neuron* 88(2): 419–31.

Chien, Hsiang Yun Sherry, and Christopher J. Honey. 2020. "Constructing and Forgetting Temporal Context in the Human Cerebral Cortex." *Neuron* 106(4): 675-686.e11. https://doi.org/10.1016/j.neuron.2020.02.013.

Dominey, Peter Ford. 2021. 17 PLoS Computational Biology *Narrative Event Segmentation in the Cortical Reservoir*. http://dx.doi.org/10.1371/journal.pcbi.1008993.

DuBrow, Sarah, and Lila Davachi. 2016. "Temporal Binding within and across Events." *Neurobiology of Learning and Memory* 134: 107–14.

Gao, Richard, Ruud L. Van den Brink, Thomas Pfeffer, and Bradley Voytek. 2020. "Neuronal Timescales Are Functionally Dynamic and Shaped by Cortical Microarchitecture." *eLife* 9: 1–44.

Glorot, Xavier, and Yoshua Bengio. 2010. "Understanding the Difficulty of Training Deep Feedforward Neural Networks." *Journal of Machine Learning Research* 9: 249–56.

Grill-Spector, Kalanit, and Kevin S. Weiner. 2013. "The Functional Architecture of the Ventral Temporal Cortex and Its Role in Categorization." *Nature Reviews Neuroscience* 23(1): 1–7.

Hasson, Uri, Janice Chen, and Christopher J. Honey. 2015. "Hierarchical Process Memory: Memory as an Integral Component of Information Processing." *Trends in Cognitive Sciences* 19(6): 304–13.

Hénaff, Olivier J., Robbe L.T. Goris, and Eero P. Simoncelli. 2019. "Perceptual Straightening of Natural Videos." *Nature Neuroscience* 22(6): 984–91.

Honey, Christopher J. et al. 2012. "Slow Cortical Dynamics and the Accumulation of Information over Long Timescales." *Neuron* 76(2): 423–34. http://dx.doi.org/10.1016/j.neuron.2012.08.011.

Honey, Christopher J., Ehren L. Newman, and Anna C. Schapiro. 2017. "Switching between Internal and External Modes: Amultiscale Learning Principle." *Network Neuroscience* 104(51): 20167–72. http://www.mitpressjournals.org/doi/abs/10.1162/NETN_a_00024.

Ilya Sutskever. 2013. "Training Recurrent Neural Networks." *Ph.D thesis*.

Iwata, Kazunori, and Naohiro Ishii. 2002. "Discrepancy as a Quality Measure for Avoiding Classification Bias." *International Syposium on Intelligent Control*: 532–37.

Jain, Shailee et al. 2020. "Interpretable Multi-Timescale Models for Predicting FMRI Responses to Continuous Natural Speech." *Advances in Neural Information Processing Systems* 2020-Decem(NeurIPS): 1–12.

Kell, Alexander J E et al. 2018. "A Task-Optimized Neural Network Replicates Human Auditory Behavior, Predicts Brain Responses, and Reveals a Cortical Processing Hierarchy." *Neuron* 98(3): 630-644.e16. http://www.ncbi.nlm.nih.gov/pubmed/29681533.

Kiebel, Stefan J, Jean Daunizeau, and Karl J Friston. 2008. "A Hierarchy of Time-Scales and the Brain." *PLoS computational biology* 4(11).

Kietzmann, Tim C. et al. 2019. "Recurrence Is Required to Capture the Representational Dynamics of the Human Visual System." *Proceedings of the National Academy of Sciences of the United States of America* 116(43): 21854–63.

LeCun, Yann, Corinna Cortes, and Christopher J C Burges. 2010. "The MNIST Database of Handwritten Digits." *URL http://yann. lecun. com/exdb/mnist* 10(34): 14.

Lee, Dong Hyun, Saizheng Zhang, Asja Fischer, and Yoshua Bengio. 2015. "Difference Target Propagation." *Lecture Notes in Computer Science (including subseries Lecture Notes in Artificial Intelligence and Lecture Notes in Bioinformatics)* 9284: 498–515.

Lillicrap, Timothy P. et al. 2020. "Backpropagation and the Brain." *Nature Reviews Neuroscience* 21(6): 335–46.

Lillicrap, Timothy P., Daniel Cownden, Douglas B. Tweed, and Colin J. Akerman. 2016. "Random Synaptic Feedback Weights Support Error Backpropagation for Deep Learning." *Nature Communications* 7: 1–10. http://dx.doi.org/10.1038/ncomms13276.

Lillicrap, Timothy P, and Adam Santoro. 2019. "Backpropagation through Time and the Brain." *Current Opinion in Neurobiology* 55: 82–89. https://linkinghub.elsevier.com/retrieve/pii/S0959438818302009.

Mahto, Shivangi, Vy A. Vo, Javier S. Turek, and Alexander G. Huth. 2020. "Multi-Timescale Representation Learning in LSTM Language Models." : 1–19. http://arxiv.org/abs/2009.12727.

Mishra, Siddhartha, and T. Konstantin Rusch. 2021. "Enhancing Accuracy of Deep Learning Algorithms by Training with Low-Discrepancy Sequences." *SIAM Journal on Numerical Analysis* 59(3): 1811–34.

Mitchell, Melanie. 2020. "On Crashing the Barrier of Meaning in Artificial Intelligence." *AI Magazine* 41(2): 86–92.

Mozer, M, R Lippmann, J Moody, and D Touretsky. 1992. "Induction of Multiscale Temporal Structure." *Advances in Neural Information Processing Systems 4*: 275–82.

Murray, John D. et al. 2014. "A Hierarchy of Intrinsic Timescales across Primate Cortex." *Nature Neuroscience* 17(12): 1661–63. http://dx.doi.org/10.1038/nn.3862.

Nayebi, Aran et al. 2018. "Task-Driven Convolutional Recurrent Models of the Visual System." *Advances in Neural Information Processing Systems* 2018-Decem(Nips): 5290–5301.

Paszke, Adam et al. 2019. "PyTorch: An Imperative Style, High-Performance Deep Learning Library." *Advances in Neural Information Processing Systems* 32(NeurIPS).

Raut, Ryan V., Abraham Z. Snyder, and Marcus E. Raichle. 2020. "Hierarchical Dynamics as a Macroscopic Organizing Principle of the Human Brain." *Proceedings of the National Academy of Sciences of the United States of America* 117(34): 20890–97.

Robert M. French. 1999. "Catastrophic Forgetting in Connectionists Networks." *Trends in Cognitive Sciences* 3(4): 128–35.

Smith, Stephen M. et al. 2013. "Functional Connectomics from Resting-State FMRI." *Trends in Cognitive Sciences* 17(12): 666–82.

Soltani, Alireza, John D. Murray, Hyojung Seo, and Daeyeol Lee. 2021. "Timescales of Cognition in the Brain." *Current Opinion in Behavioral Sciences* 41: 30–37. https://doi.org/10.1016/j.cobeha.2021.03.003.

Spoerer, Courtney J, Patrick McClure, and Nikolaus Kriegeskorte. 2017. "Recurrent Convolutional Neural Networks: A Better Model of Biological Object Recognition." *Frontiers in psychology* 8: 1551. www.frontiersin.org.

Tieleman, T., and Geoffrey Hinton. 2012. "Divide the Gradient by a Running Average of Its Recent Magnitude." *COURSERA: Neural Networks for Machine Learning - RMSprop lecture* 04(05): 107–11.

Ulanovsky, Nachum, Liora Las, Dina Farkas, and Israel Nelken. 2004. "Multiple Time Scales of Adaptation in Auditory Cortex Neurons." *Journal of Neuroscience* 24(46): 10440–53.

Wang, Jianfeng, and Xiaolin Hu. 2021. "Convolutional Neural Networks with Gated Recurrent Connections." *IEEE Transactions on Pattern Analysis and Machine Intelligence*: 1–16.

Whittington, James C R, and Rafal Bogacz. 2019. "Theories of Error Back-Propagation in the Brain." 23(3): 235–50.

Wiskott, Laurenz, and Terrence J Sejnowski. 2002. "Slow Feature Analysis : Unsupervised Learning of

Invariances.” 770: 715–70.
Xiao, Han, Kashif Rasul, and Roland Vollgraf. 2017. “Fashion-MNIST: A Novel Image Dataset for Benchmarking Machine Learning Algorithms.” : 1–6. http://arxiv.org/abs/1708.07747.
Yamins, Daniel L. K. et al. 2014. “Performance-Optimized Hierarchical Models Predict Neural Responses in Higher Visual Cortex.” *Proceedings of the National Academy of Sciences* 47(3): 237–46.
Yamins, Daniel L K, and James J. DiCarlo. 2016. “Using Goal-Driven Deep Learning Models to Understand Sensory Cortex.” *Nature neuroscience* 19(3): 356–65. http://www.ncbi.nlm.nih.gov/pubmed/26906502.

# Supporting information

## S1 Appendix.

### Testing the efficacy of linear recurrence and gating mechanisms in categorizing noisy data

One natural benefit of having recurrence in the network is that accumulating information over autocorrelated data that contains uncorrelated noise, reduces the overall noise in the integrated signal. Therefore, when facing a noisy test data, the performance accuracy of the model equipped with linear recurrence and gating would be less affected than memoryless model. We tested the classification accuracy of networks with and without memory on noisy MNIST. We found that compared to the memoryless model trained with randomly ordered data, the performance of model with linear recurrence and gating trained with autocorrelated data was significantly less affected by the noise.

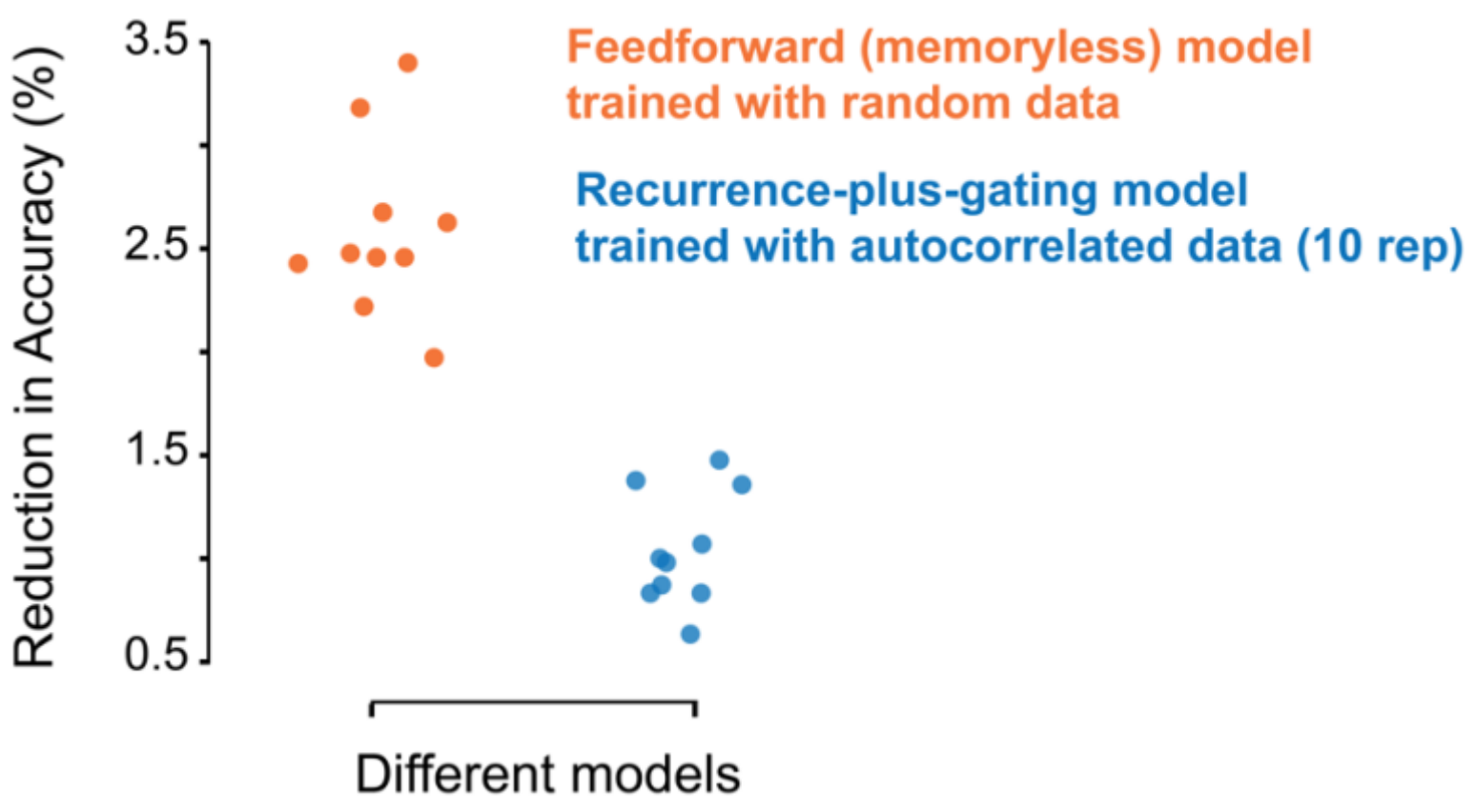

**S Fig. 1. Change in classification accuracy when recurrent and feedforward models are tested on noisy data.**

Y-axis shows the percentage of reduction in classification accuracy when tested on noisy data. X-axis shows different models. Each dot shows the reduction in classification accuracy for each single run with a different seed number (different initialization).

## S2 Appendix.

### How is the accuracy of recurrent networks affected by accumulating information across sequentially correlated training samples?

We hypothesized that accumulating information over multiple samples from the same category with overlapping features in recurrent networks enhances decision accuracy. We examined this hypothesis by evaluating the frequency of misclassifying an item as a function of its position within an autocorrelated data stream. In particular, we consider the case in which 10 items from the same category are presented sequentially, and we examine performance for items located from position 1 to position 10 within the 10-sample sequence. We found that most misclassifications happened in the first position (right after the switch to a new category. As information was integrated across the autocorrelated training samples, however, the misclassification frequency significantly dropped. We used the same neural architecture and hyperparameters as those explained in Section 2.

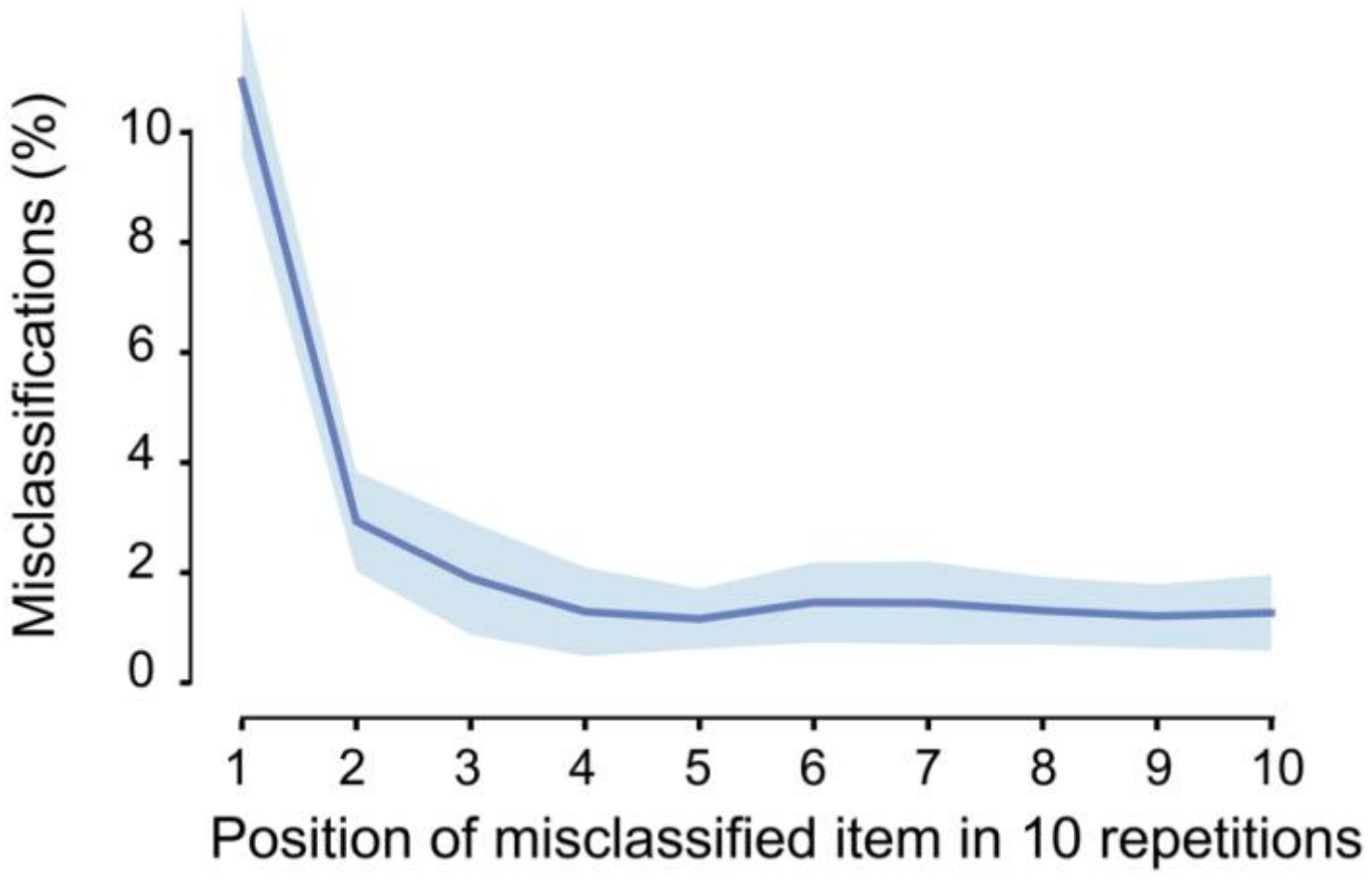

**S Fig. 2. Percentage of misclassifications for each position in 10 repetitions. (MNIST dataset)**

Y-axis shows the percentage of misclassification. X-axis shows the position of the tested iteration in 10-repetiiton of one category.

## S3 Appendix.

### Effects of temporal autocorrelation on category-Learning in an LSTM trained with backpropagation through time

We explored how training with temporally correlated data affected category-learning in an LSTM trained with backpropagation through time. The dataset, network dimension, learning rate, and optimization method were identical to ones used in models described in Fig 1. We found that increased autocorrelation in the training data resulted in better classification performance in the LSTM. This suggests that both the LSTM trained with BPTT and the linear-recurrence-plus-gating model are similarly affected by the autocorrelation in training data, since learning in both models is improved with increased autocorrelation in the training data.

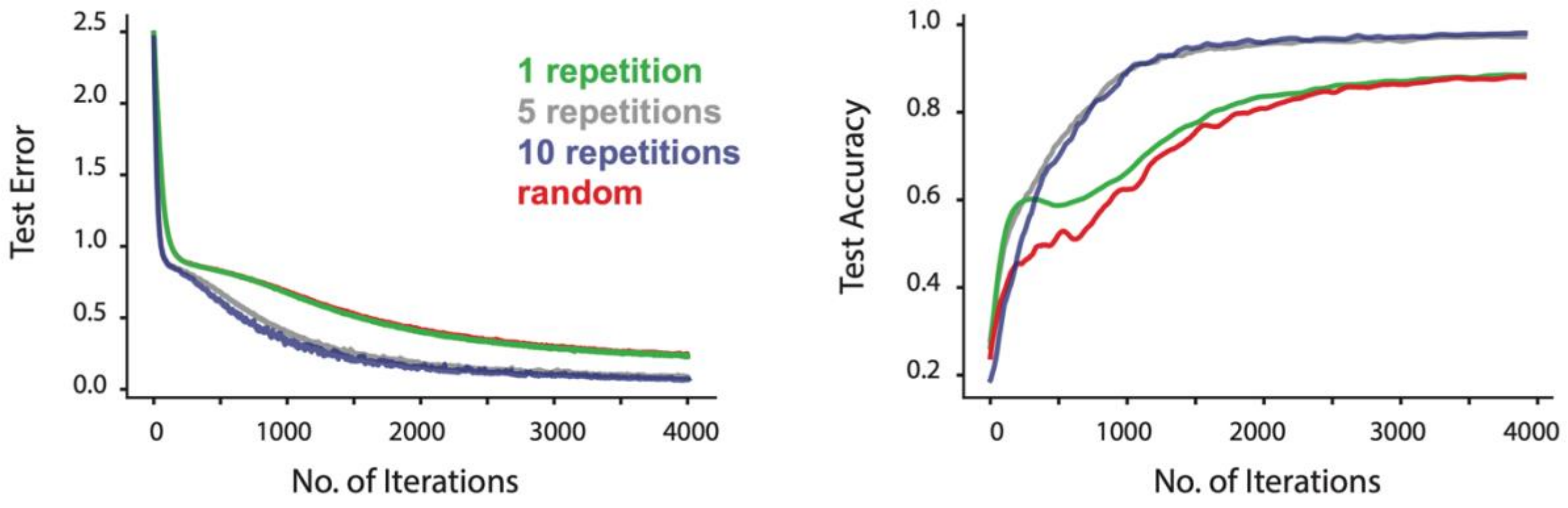

**S Fig. 3. Effects of autocorrelation in data on category learning in LSTM.**

Left) Test error of an LSTM model across different amounts of autocorrelation in training data. Right) Test classification accuracy of an LSTM model across different amounts of autocorrelation in training data.

## S4 Appendix.

### Comparing an LSTM with the linear-recurrence-plus-gating model

We compared LSTM, the linear-recurrence-plus-gating model, and the feedforward memoryless model in their category learning ability. We tested the LSTM and the linear-recurrence-plus-gating model on the same data structure that they were trained on (e.g. training with 5-repetitions and testing on 5-repetitions). Thus, we included memory in the testing process and we tested the models on the same stimulus-order that they were trained on. In this setting, where no generalization was required, the LSTM was the most efficient learner, and the linear-recurrence-plus-gating model was second-best, superior to the memoryless model.

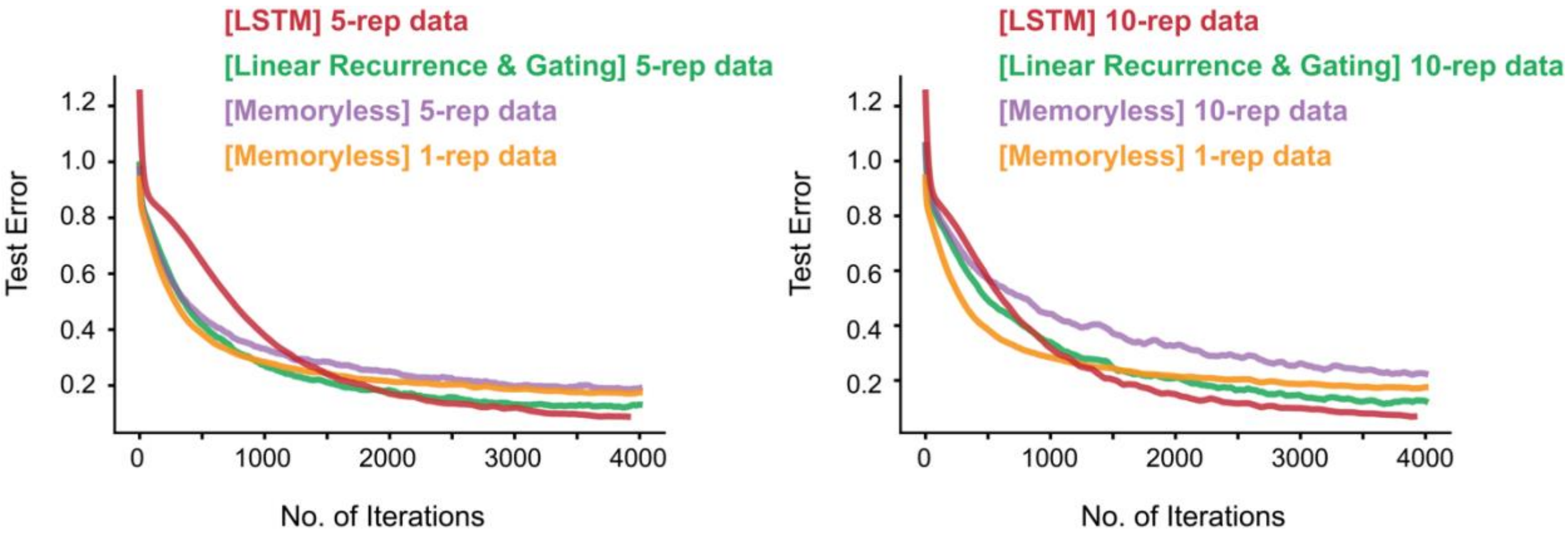

**S Fig. 4. Comparing the LSTM model, linear-recurrence-plus-gating model, and the memoryless model.**

Left) Test error for classification with autocorrelation level equal to 5-repetitions of each category. Right) Test error for classification with autocorrelation level equal to 10-repetitions of each category. In both plots, we also provide an orange reference curve [memoryless with 1-repetition], that illustrates the performance of a model trained with minimum autocorrelation.

## S5 Appendix.

### Comparing the linear recurrence approach against mini-batch training

Mini-batching and linear recurrence are conceptually similar, because in one case the gradients are averaged across training samples, and in the other case the activation patterns themselves are averaged (weighted average). However, the two phenomena are not equivalent.

First, we showed that batching and linear-recurrence-plus-gating have effects on learning that are qualitatively different as we vary the autocorrelation of the data (S Fig. 5). Batching of size 10 with 10-rep data leads to less efficient learning than batching of size 1 with 1-rep data. In contrast, in the linear-recurrence-plus-gating model, 10-rep leads to more efficient learning than 1-rep. Moreover, in batching, the outcome of different batch sizes eventually converged to the same level of performance. However, in the linear-recurrence-plus-gating model, increasing the repetitions (increased smoothness) resulted in improved performance and this advantage persisted over time.

Note that the x-axis in these figures shows the number of samples seen by the models, rather than the number of training iterations. Therefore, for the same number of samples, there are fewer updating iterations in mini-batching, and thus a shorter run time. This is one of the main reasons for the desirability of mini-batch training. However, given the same number of training samples, the weighted-averaging of hidden states in the linear-recurrence-plus-gating case results in a lower generalization error compared to averaging gradients in the mini-batching (green curves are identical across left and right plots). Having said that, the purpose of this comparison is not to demonstrate the advantage of linear-recurrence-plus-gating over mini-batching. Instead, we hope this analysis show that mini-batching and linear recurrence are qualitatively different and serve different purposes.

Second, when our model was trained on a data structure in which consecutive samples did not share local features, recurrence advantage was eliminated (Fig 2). In contrast, mini- batching would work well in this setting.

To see the difference between linear recurrence and batching directly, consider a toy example. For batching at size 6, the sequence of training samples ABABAB and AAABBB are exactly equivalent, producing the identical gradient. However, in the case of linear recurrence, the sample sequence AAABBB and ABABAB can generate totally different gradient sequences. For example, in the case where A = -B, there will be dramatic interference between consecutive stimuli in the ABABAB case, but much less interference in the AAABBB case.

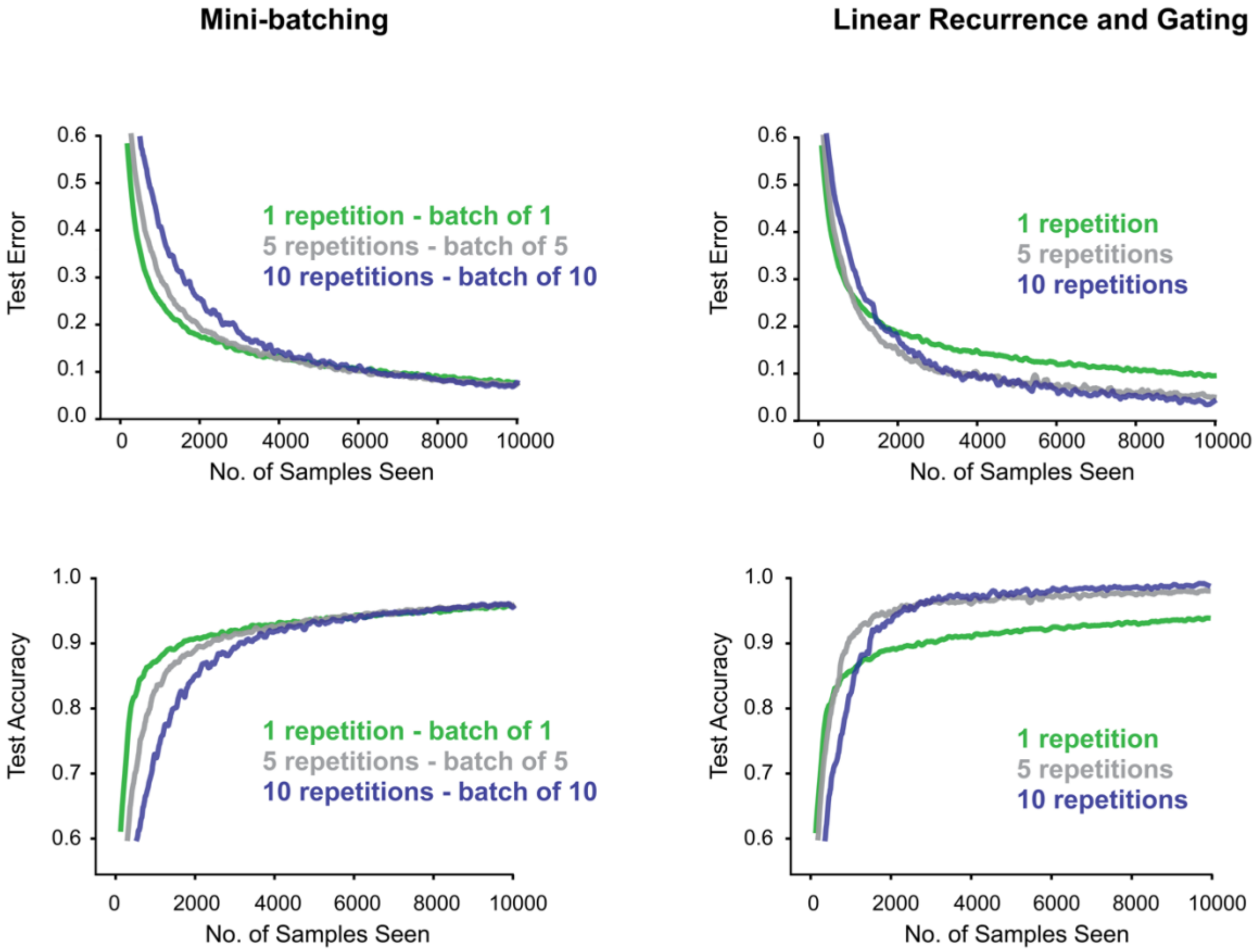

**S Fig. 5. Comparing the effects of temporal autocorrelation in data on mini-batching versus on models with linear recurrence and gating.**

Left) Test error and test accuracy of mini-batch training on MNIST data for different batch size and different level of autocorrelation in data. Right) Test error and test accuracy of linear-recurrence-plus-gating model on MNIST data for different level of autocorrelation in data. Both models had the dimension of (784, 392, 10), learning rate of 0.01. Optimization method in linear-recurrence-plus-gating model was SGD. Each curve shows the average of 5 runs with 5 different initialization seeds.

## S6 Appendix.

### Why does the memoryless AE model achieve lower test error than AE models with recurrence and gating?

The reason for the slight advantage of the no-memory model over the multi-scale model is that the multi-scale model faces a challenge of using slowly-changing internal representations to reconstruct quickly-changing features. Because all nodes in the hidden layer of the multi-scale model project to all nodes in the reconstruction later, the slowly changing hidden states of the model (i.e. the nodes with longer memory) are contributing to the reconstruction of quickly-changing features in the data stream. There is a (small) cost in the overall test error, because slowly-changing internal states are ineffective for reconstructing quickly-changing features. We emphasize that the quantity of noise introduced is small, and that it is accompanied by a significant benefit in learning more interpretable, un-mixed representations of a multiscale data stream. To demonstrate that these slow units are indeed the source of poorer learning, we tested the hypothesis that (i) a higher correlation between hidden units with memory (units with short or long memory) and fast-changing part of the output would result in a worse performance in reconstructing the fast-changing feature; whereas (ii) a higher correlation between hidden unit with no-memory and fast-changing part of the output would not result in worse performance in reconstructing the fast-changing feature. These hypotheses were confirmed in our analyses.

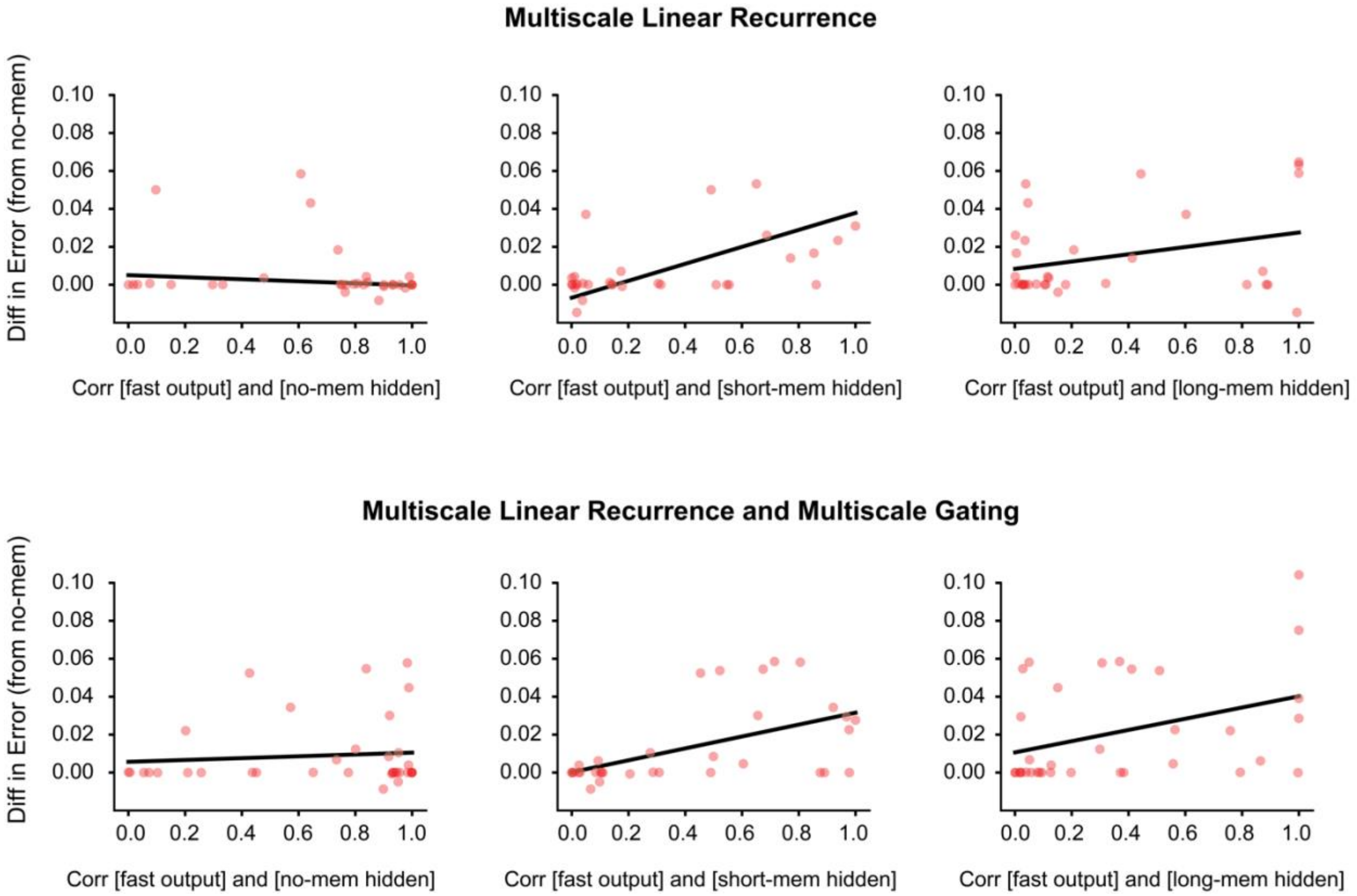

**S Fig. 6. Relationship between correlation of hidden nodes with fast output, and the reconstruction error for fast output.**

Top) multiscale linear recurrence (M-LR) model: y-axis shows the difference between test reconstruction error for fast feature (e.g. error of reconstructing fast feature in M-LR model - error of reconstructing fast feature in memoryless model); x-axis in left plot shows the correlation of fast output with no-memory hidden nodes, x-axis in middle plot shows the correlation of fast output with short-memory hidden nodes, x-axis in right plot shows the correlation of fast output with long-memory hidden nodes. Bottom) Plots show similar results shown at the top row for the model with multiscale linear recurrence and multiscale gating (M-LR-M-G).

## S7 Appendix.

### How does the magnitude of memory coefficients interact with efficiency and timescale selectivity of the AE models?

Here, we tested how for each AE architecture, the magnitude of memory coeffects interact with the timescale selectivity and the efficiency of the models. We hypothesized that equipping AE models with recurrence and gating enables them to learn temporally-interpretable representations, however, that benefit comes with a small increase in the overall test error. Consistent with our prediction, we found that in all AE models, higher memory coefficients showed a higher increase in the test error. In AE models with recurrence that did not have the gating mechanism, higher memory coefficients showed worse performance: higher memory coefficients were correlated with higher test error and lower timescale selectivity (see “uniform memory” and “multiscale memory” in S Fig. 7). In contrast, in AE models with recurrence and multiscale gating, higher memory coefficients were correlated with a smaller increase in the test error but also an increase in timescale selectivity (see “uniform memory & multiscale gating” and “multiscale memory and multiscale gating” in S Fig. 7). Therefore, although the gating mechanism allows the AE models to selectively combine related information and prevent interference from unrelated information, higher memory coefficients do not necessarily lead us to a *better* model. Thus, in practical settings, the memory coefficients may be adjusted to trade off timescale selectivity (when we want to learn separated representations for distinct sources) and test error (when we wish to optimize the accuracy of reconstruction of momentary input).

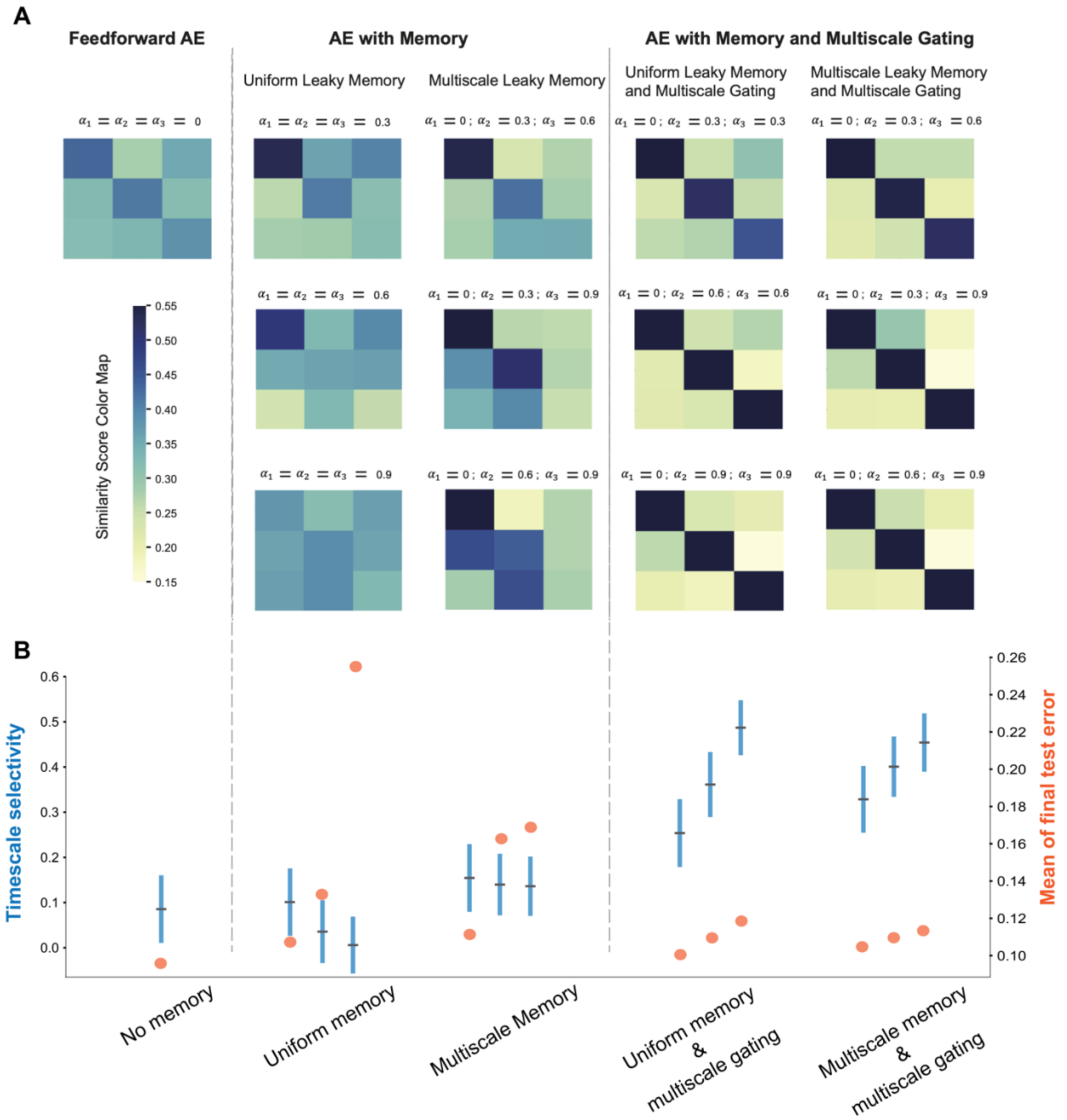

**S Fig. 7. Timescale selectivity and final test error for different AE models.**

A) Results of similarity matrices for 3 groups of models: feedforward AE, AE with memory, and AE with memory and multiscale gating. Each similarity matrix shows the similarity pattern for a model with a specific set of memory coefficients shown above it. B) The colored bars show the timescale selectivity. Error bars show the 95% confidence interval for 40 runs with different ransom initializations. The gray dots show the average of final test error for 40 runs. For each group of models (e.g. uniform memory, etc.), we tested 3 models with 3 sets of memory coefficients. From left to right, memory coefficients increase gradually. For instance, for uniform memory, memory coeffects of all the internal nodes in the models left model were 0.3, for the middle model were 0.6, and for the right model were 0.9.

## S8 Appendix.

### Unsupervised local resetting mechanism

For learning multiscale data in autoencoders, we implemented a simple "resetting" mechanism that was unsupervised way employed only local computations. Our resetting criterion compares the absolute difference between two consecutive inputs against their absolute mean:

$$| I_t - I_{t-1} | > | ( I_t + I_{t-1} ) /2 | \qquad \textbf{S Eq. 1}$$

where $t$ represents the iteration number during training, $I_t$ is the current state, $I_{t-1}$ previous state.
This method is consistent with neurophysiological studies that demonstrate a sudden shift in memory representations in the face of a surprise in the input stimuli (Chien and Honey 2020; DuBrow and Davachi 2016).

For multi-scale resetting, the distinct reset mechanisms are responsive to different input features: the slow reset is based on the amount of change in the slowly-changing input features, the medium reset is based on the amount of change in the medium-changing input features, and the fast reset is based on the amount of change in the fast-changing input feature.

The method we used is somewhat artificial, as it depends on knowing which feature changes quickly or slowly, but an equivalent outcome can be achieved by smoothing the inputs to the reset mechanism to make them sensitive to different scales. For example, to generate a reset mechanism that is sensitive to slow changes, the difference computation in the numerator can be a difference between the current input and a running-average over the past 5 inputs.